\documentclass[acmsmall]{acmart}
\usepackage{graphicx}
\usepackage{subcaption}
\usepackage{balance}
\usepackage{tabularx}
\usepackage{multirow}

\AtBeginDocument{%
  }

\setcopyright{acmlicensed}
\copyrightyear{2025}
\acmYear{2025}
\acmDOI{XXXXXXX.XXXXXXX}

\acmJournal{TIST} 

\acmVolume{1}
\acmNumber{1}
\acmArticle{1}
\acmMonth{1}

\begin{document}

\title{From Numbers to Prompts: A Cognitive Symbolic Transition Mechanism for Lightweight Time-Series Forecasting}

\author{Namkyung Yoon}
\email{nkyoon93@korea.ac.kr}
\orcid{0000-0002-2144-6664}
\affiliation{%
  \institution{School of Electrical Engineering, Korea University}
  \city{Seoul}
  \country{Republic of Korea}
}

\author{Hwangnam Kim}
\email{hnkim@korea.ac.kr}
\orcid{0000-0003-4322-8518}
\affiliation{%
  \institution{School of Electrical Engineering, Korea University}
  \city{Seoul}
  \country{Republic of Korea}
}
\authornote{Corresponding author}

\renewcommand{\shortauthors}{Yoon and Kim}


\begin{abstract}
Large language models have achieved remarkable success in time series prediction tasks, but their substantial computational and memory requirements limit deployment on lightweight platforms.
In this paper, we propose the Symbolic Transition Mechanism (STM)—a novel framework that bridges numeric time-series data and language models through symbolic abstraction and prompt engineering. 
STM transforms continuous time series values into symbol tokens with quantization techniques based on human cognitive structures, and captures temporal dynamics through structured transformations of symbols, enabling fast engineering-based predictions in which language models focus on critical parts of time series data.
STM is a general-purpose mechanisms that ensure the integrity of backbone language models, but they significantly improve their efficiency by inferring the dynamic and structured patterns inherent in time series data.
We evaluated STM on various time series datasets, paired with four small language models (SLM) with limited computational environments.
For all models, STM achieves error reductions of up to 69\% in MAE and 90\% in MSE compared to the default backbone SLM without STM.
These results demonstrate the potential of STM as an efficient, adaptable layer for symbol-driven time series prediction using foundation models. The accuracy improvements were made at negligible resource costs, with maximum GPU memory of the base model increasing by approximately 0.06\% and latency overhead increasing by only 0.64\%.

\end{abstract}



\maketitle

\section{Introduction}

{T}{ime} series forecasting is a critical task in numerous real-world applications, ranging from climate and weather prediction to industrial Internet of Things (IoT) systems \cite{lim2021time, ren2023deep, kashpruk2023time}. Recent studies have shown that Large Language Models (LLMs) can effectively capture temporal patterns by transforming numerical time series into textual prompts, allowing the models to leverage their powerful contextual understanding for prediction tasks such as Time-LLM\cite{jin2023time}.

However, the computational overhead and memory requirements of LLM often limit their practicality in resource-constrained edge environments, where low latency and low power operations are required \cite{wolters2024memory}.

Building on this trend, there has been growing interest in adopting Small Language Models (SLMs) as lightweight yet high-performance solutions for edge environments\cite{sangwan2025cloud,sarhaddi2025llms}.

However, SLMs are still suboptimal for effectively capturing complex temporal dependencies with relatively small capacity compared to LLMs, although they have low computational cost and memory consumption \cite{wang2024comprehensive}.
In addition, traditional self-attention mechanisms, which are widely used these days, are computationally intensive and can lead to scaling problems in embedded IoT or edge scenarios \cite{yoon2024detecting}.

To bridge this gap, we propose an efficient and interpretable novel symbolic transition mechanism (STM) for time series prediction.
The proposed STM is designed to complement and enhance existing LLM-based prediction frameworks.
While these frameworks introduce an effective prompt pipeline for numerical time series, their performance essentially depends on how clearly the time structure can be conveyed to the language model.
STM improves representations that are not explicitly represented from raw number prompts injected in the conventional way.

Our STM proposed in this paper consists of a core three-step strategy as follows:
\begin{enumerate} \item \textit{Symbolic Encoding of Time Series Data:} The proposed STM mechanism quantizes it with discrete symbols based on cognitive psychology to facilitate encoding on continuous time series data such as temperature or traffic and reduce numerical complexity.

\item \textit{Transition-based Weighting:} By allocating distance-based weights between symbols, the STM mechanism identifies significant changes and gives significance to larger transitions. This effectively highlights significant changes in time series data, such as abrupt temperature fluctuations.

\item \textit{Pattern Periodicity Detection:} Based on the uniqueness of the periodic pattern implied by the time series data, the proposed STM detects segments of the iterative pattern and gives additional weight.

\end{enumerate}

This makes STM a generalizable framework that can be integrated with a variety of commercial SLMs, allowing for the prediction of improvements in time series prediction capabilities without huge computational costs.

Our contributions are summarized as follows:
\begin{itemize}
  \item We propose STM, a symbolic augmentation module that dramatically improves temporal transition, periodic pattern, and symbolic dynamics modeling without modifying the backbone parameters of language models.

  \item We find that the five-level symbolic quantization of STM provides an effective-balance between granularity and generalization in SLMs, showing an intriguing alignment with classical findings on human cognitive capacity such as Miller's law.

  \item Through extensive experiments across four SLMs and two datasets, 
  STM consistently reduces MAE and MSE with negligible overhead in GPU memory and inference latency, demonstrating its practicality for edge-oriented deployments.
\end{itemize}

The rest of this paper is organized as follows. Section 2 is a brief overview of the relevant research on the use of language models and attention mechanisms for prior knowledge of STM that we propose. Section 3 describes the details of the proposed STM and how it is integrated with SLM. Section 4 continues the discussion of experimental results and insights using time series data. Finally, Section 5 concludes the paper and outlines potential directions for future research.

\section{Preliminary}
\label{preliminary}
In this section, we present the main concepts that underlie the proposed approach. First, we provide a basic overview of AI-based language models that have recently become a hot topic in various fields.

Subsequently, we discuss the principles of attention mechanisms, summarize the role of attention mechanisms in capturing long-range dependencies in sequential time series data mathematically, and explain how they extend to the proposed STM.

\subsection{Language Model}
Language models (LMs) aim to model human-used natural languages by learning probabilistic distributions over tokenized sequences~\cite{rajaraman2024toward}. Starting from sequence-to-sequence models, the introduction of attention mechanisms led to the Transformer architecture~\cite{vaswani2017attention}, which enabled more effective modeling of long-range dependencies.

Recent LMs, including GPT, LLaMA, Phi, DeepSeek, and Gemma, are large transformer models trained to generate coherent sequences aligned with real world tasks~\cite{shao2024survey}. These models predict the likelihood of tokens based on preceding context, supporting tasks such as translation and question answering.

As language models are now increasingly applied to domains beyond natural language, including time series prediction, methods for converting numerical data into text prompts have gained attention~\cite{zhang2024systematic}.

\subsubsection{Small Language Model}
With the recent development of various techniques of artificial neural networks such as deep learning, large language models (LLM) have been actively studied for various tasks in natural language processing~\cite{khan2023exploring}. However, LLM represents convenience and efficiency in various fields, but the vast amount of pre-training processes and parameters consume a lot of resources in the computational process~\cite{bai2024beyond}.
For these limitations, the study of small-language models (SLM) that maintain as many modeling capabilities as possible while reducing the computational resources of LLM is also being conducted in many ways~\cite{jovanovic2024compacting}.

SLM is a language model that has parameters ranging from 1 to dozens of billion and requires 4GB to 8GB of RAM to operate with less computational power than LLM, aimed at resource-limited environments or private uses, such as mobile devices and edge computing~\cite{popov2024overview, mekala2024smaller}.
SLM aims to reduce the number of parameters and reduce memory usage by applying techniques such as knowledge distillation, model pruning, and quantization to LLM depending on the task~\cite{zhu2024survey}.

This makes SLM more favorable for edge distribution and real-time inference than LLM, making it suitable for resource-constrained environments such as IoT devices and mobile applications~\cite{zhang2025rise}.
However, SLM with reduced parameters can be difficult to capture contextual information about data with larger data capacities or complex time dependencies~\cite{matarazzo2025survey}.
These limitations can be fatal for time series tasks that require subtle pattern recognition~\cite{lin2012pattern}.
Therefore, we propose a novel attention mechanism to achieve high accuracy without compromising the efficiency of SLM.

Classical transformer-based forecasting models such as Informer, Autoformer, and TFT introduced architectural inductive biases for long-range temporal modeling~\cite{zhou2021informer, wu2021autoformer, lim2021temporal}.
Recently, with the advent of LM, research has gone beyond the performance of existing architectures.
In particular, \textit{Time-LLM}~\cite{jin2023time} demonstrated that converting numerical sequences into text prompts enables pretrained LLMs to perform time series forecasting through \emph{prompt reprogramming}.
This approach leverages the inherent contextual reasoning abilities of LLMs and has shown strong performance across domains~\cite{jiang2024empowering}.

However, LLM-based forecasting relies on large Transformer architectures with memory-intensive QKV attention, making them impractical for deployment in resource-constrained environments~\cite{zhu2019empirical, dao2022flashattention}. 
Moreover, while LLMs excel at contextual reasoning, they do not explicitly embed the inductive biases toward temporal structure emphasized in earlier time series models.

Motivated by these limitations, we propose a lightweight symbolic attention mechanism that injects temporal inductive bias directly into SLMs. 

Previous studies such as Time-LLM have shown that language models can perform surprisingly well in time series prediction when inputting numerical sequences through carefully designed prompts.
STM is compatible and complementary to these approaches. Instead of changing the prompt mechanism, we augment STM by adding symbolic time patterns that are not explicitly encoded to the raw numerical input.
Thus, STM can be viewed as a plug-in extension that improves the temporal resolution and inference depth of prompt-based LLM or SLM forecasters.

Our STM enhances time series forecasting accuracy without modifying model architecture or requiring retraining, making it widely applicable across different SLMs.

\subsection{Attention mechanism}
Attention mechanisms are pivotal in modern AI, particularly in LLM such as transformer-based architectures~\cite{lin2024infinite}. Attention allows models to dynamically focus on relevant parts of the input, enhancing their ability to handle long-range dependencies as follows:
\begin{equation}
\text{Attention}(Q, K, V) = \text{softmax}\left(\frac{QK^T}{\sqrt{d_k}}\right)V,
\label{a}
\end{equation}
where \(Q\) (Query), \(K\) (Key), and \(V\) (Value) are matrices representing the input embeddings, and \(d_k\) is the dimensionality of the key vectors. This formulation enables the model to effectively compute the relevance scores between the input elements, ensuring that the most pertinent symbols receive a higher weight in the final representation~\cite{vaswani2017attention}.
In this study, we do not modify the internal QKV attention of small language models.
Instead, STM operates as a lightweight auxiliary module that extracts symbolic representations, transition strengths, and periodic cues from time series data, and injects them into the prompt space.
This external symbolic structure highlights important temporal transitions without replacing or altering the underlying attention mechanism of the model.
 
\begin{figure*}
\centering \makeatletter\IfFileExists{fig/fig2.png}{\includegraphics[width=0.95\textwidth]{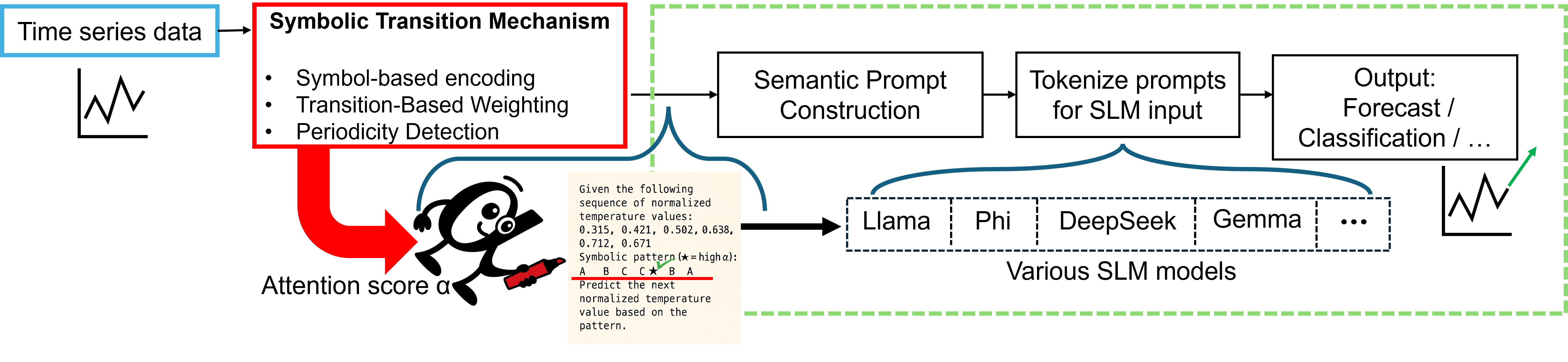}}{}
\makeatother 
\caption{{Overview of the proposed time series prediction model with the Symbolic Transition Mechanism.}}
\Description{Overview of the proposed time series prediction model with the Symbolic Transition Mechanism.}
\label{overview}
\end{figure*}

\section{Design Principle and Approach}
\label{design_principle_approach}

In this section, we propose STM, a novel attention mechanism for working language models that identify and predict characteristics based on the periodicity of time series data.
STM aims to enhance the overall performance of the underlying backbone model by integrating various SLMs, each applying the time-series prediction reprogramming described in Section 2.

Furthermore, STM induces time series data into text, allowing language models to understand the underlying patterns and interdependencies within time series data.
With this, STM aims to be easily integrated into various language model-based pipelines, such as finetuned LLM and LLM agents, as well as the reprogramming techniques of Time-LLM. Therefore, it acts as a universal inductive bias mechanism to improve time series understanding across architectures.

\subsection{Theoretical Approach}

We approach the main properties of time series data on a theoretical basis, identifying seasonality, trend, and level shifts as three fundamental characteristics inherently embedded in time series.
These properties arise naturally from the dynamic and structured nature of time series data, and serve as fundamental motivations for STM design:

\begin{itemize}
    \item \textbf{Trend:} Time series data are constantly moving and fluctuating in value over time, and the STM encodes the direction and size of the trend information.

    \item \textbf{Level Shifts:} Level shifts are sudden changes in the mean or variance of time-series data that occur suddenly and can occur at any time. These changes suggest a structural transition and thus allow them to be captured by the STM.
    
    \item \textbf{Seasonality:} Time series often contain patterns that are repeated by periodic factors over fixed intervals. To capture this, STM highlights regular time structures by identifying recurring segments in time series data and assigning periodic weights.
\end{itemize}

These characteristics have long been recognized in the time series forecasting literature as crucial for effective modeling \cite{crone2010feature}.
Based on this, the STM to further improve the time series prediction performance of SLMs consists of three components, as shown in Fig.~\ref{overview}, where each corresponds directly to a core theoretical property of time series data: symbol levels capture trend, transition magnitudes reflect level shifts, and periodicity detection models seasonal repetition.

\begin{enumerate}
    \item \textbf{Symbol-Based Encoding:} Continuous time series values are quantized into discrete symbols to reduce numerical complexity and enable interpretable pattern analysis.
    \item \textbf{Transition-Based Weighting:} Differences between consecutive symbols are emphasized, as abrupt transitions often carry critical information for forecasting tasks.
    \item \textbf{Periodicity Detection:} Repeating temporal patterns are identified and assigned greater importance, reflecting their contribution to long-term predictive performance.
\end{enumerate}

\begin{figure}
\centering \makeatletter\IfFileExists{fig/stm2.png}{\includegraphics[width=0.9\textwidth]{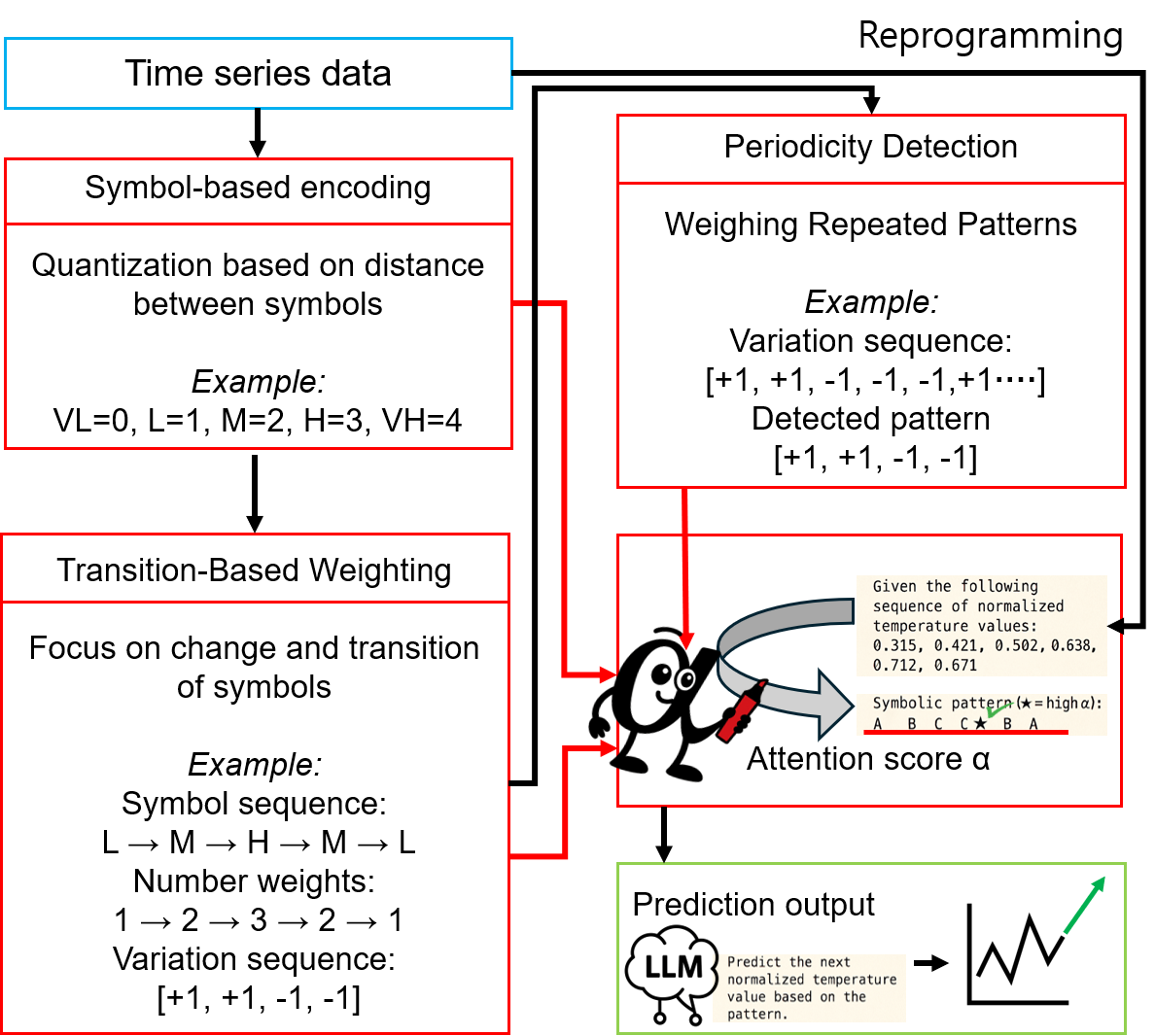}}{}
\makeatother 
\caption{{The structure of Symbolic Transition Mechanism (Red box).}}
\Description{{The structure of Symbolic Transition Mechanism (Red box).}}
\label{stm}
\end{figure}

\subsection{Symbolic Transition Mechanism}
\label{symbol}
This subsection details each step of the STM for properly learning and predicting the theoretical properties inherent in time series data, as shown in Fig.~\ref{stm}. 
STM reprograms raw numerical sequences into symbolic patterns, transition variations, and periodicity-aware representations that can be effectively interpreted by lightweight language models.

\begin{figure*}[t]
    \begin{subfigure}[t]{0.485\textwidth}
        \includegraphics[width=\textwidth]{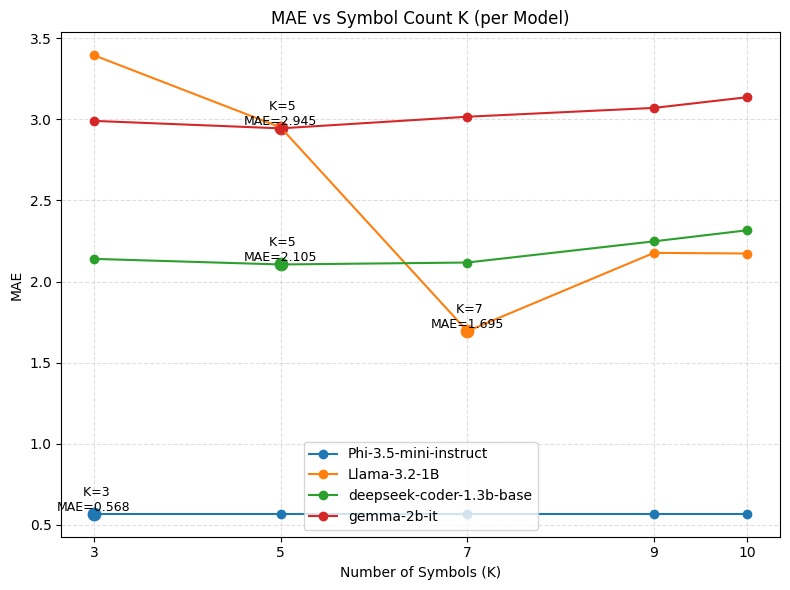}
        \caption{MAE trends across symbol counts \(K\).}
        \Description{MAE trends across symbol counts \(K\).}
    \end{subfigure}
    \hfill
    \begin{subfigure}[t]{0.485\textwidth}
        \includegraphics[width=\textwidth]{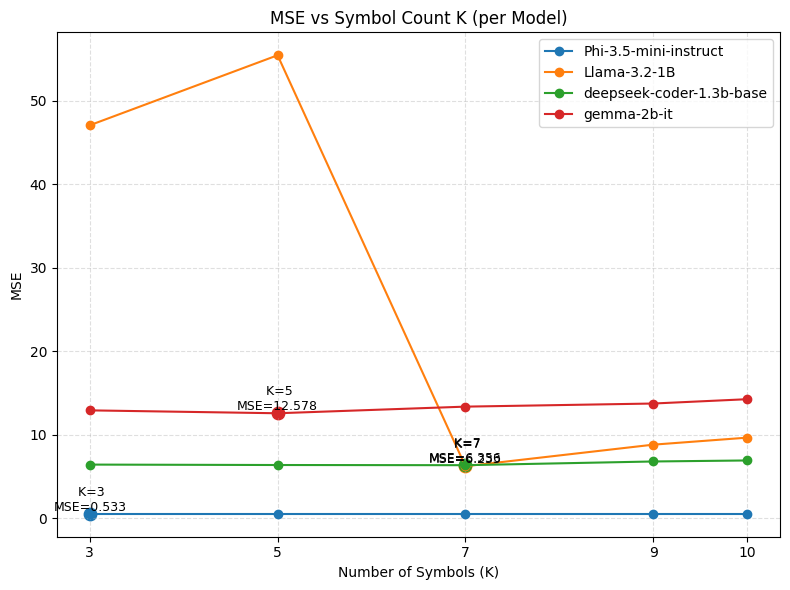}
        \caption{MSE trends across symbol counts \(K\).}
        \Description{MSE trends across symbol counts \(K\).}
    \end{subfigure}

    \caption{Validation with symbol ablation across four SLMs. Five-level encoding consistently provides optimal or near-optimal performance across models.}
    \Description{Validation with symbol ablation across four SLMs. Five-level encoding consistently provides optimal or near-optimal performance across models.}
    \label{symbol_ablation}
\end{figure*}

\subsubsection{Symbol-Based Encoding}

STM discretizes continuous time series values into symbolic representations by dividing their value range into uniformly spaced intervals. 
Each interval is assigned a distinct symbol, enabling the transformation of numerical sequences into interpretable patterns that can be directly injected as textual cues into a prompt. 
This symbolic encoding does not aim for compression; rather, it emphasizes interpretability and structural consistency across time, aligning with the reprogramming pipeline illustrated in Fig.~\ref{stm}.

In this work, we adopt five symbolic levels:
\[
\{\texttt{VL}=A,\, \texttt{L}=B,\, \texttt{M}=C,\, \texttt{H}=D,\, \texttt{VH}=E\}.
\]
The rationale for selecting five symbolic levels in the STM framework is grounded in Miller’s law from cognitive psychology, which states that humans are unable to segment continuous stimuli with infinite precision~\cite{miller1956magical}. 
Miller’s law further suggests that information presented within an optimal load range of 5–9 items minimizes cognitive effort, enabling users to process and remember information more efficiently~\cite{kalyuga2011informing}.   
Human memory–based information processing is also known to retain approximately \(7 \pm 2\) meaningful units, regardless of the specific numerical modality.

From this classical cognitive limitation, we draw inspiration that language models—trained predominantly on human-generated text—likewise tend to compress continuous variations into a small number of qualitative symbolic categories\cite{wang2023aligning}.

To validate this hypothesis, we perform a symbol-count ablation by varying the number of quantization levels \(K \in \{3, 5, 7, 9, 10\}\) across four representative SLMs: Phi-3.5-mini-instruct, Llama-3.2-1B, DeepSeek-Coder-1.3B, and Gemma-2B-it.  
We demonstrate the effectiveness of STM using commonly used metrics in time series prediction tasks: mean absolute error (MAE), mean square error (MSE), and its percentage improvement \cite{reich2016case}.
For each model-K pair, the predictor generated next-step predictions by randomly selecting for the test samples used in Section 4, resulting in MAE and MSE metrics calculated after inverse normalization to evaluate their sensitivity to symbolic granularity.

The results, summarized in Fig.~\ref{symbol_ablation}, show consistent trends in MAE and MSE: It shows consistent trends in MAE and MSE: Overall, $K=5$ offers the best or second best performance across models, while Phi-3.5-mini is significantly invariant to $K$ selection.

\begin{itemize}
    \item \(K=5\) yields the best or near-best performance in DeepSeek-Coder, Gemma-2B-it models.
    \item Llama-3.2-1B achieves its minimum MAE at \(K=7\), but \(K=5\) remains the second-best and more stable choice.
    \item Phi-3.5-mini-instruct indicates no sensitivity to \(K\), giving the same prediction for all tested \(K\) values, indicating a stable internal abstraction regardless of sign.
\end{itemize}

These results provide empirical grounding that five symbolic levels form a cognitively and computationally efficient balance between granularity and generalization.
In all subsequent experiments in Section 4.2–4.3, we fixed K=5, the number of symbols representing the most stable performance, and we propose that symbols of \(7 \pm 2\) can be used as symbols for various language models.
This suggests similarities to human cognitive abilities in Miller's law, which categorizes changes in information into a small number of perceptual clumps.

This symbolic encoding reduces numerical complexity, improves robustness under noise, and produces human-interpretable patterns—such as
\texttt{AABCCCDDEE}—that can be directly incorporated into prompts.

We additionally define a distance function \(d\colon \mathcal{S}^2 \to \mathbb{R}\), 
where \(\mathcal{S} = \{\texttt{VL}, \texttt{L}, \texttt{M}, \texttt{H}, \texttt{VH}\}\):
\begin{equation}
d(\texttt{X}, \texttt{Y}) = \left| w(\texttt{X}) - w(\texttt{Y}) \right|,
\label{d}
\end{equation}
where the weight function \(w(\cdot)\) maps each symbol to an ordinal scale in \(\mathbb{Z}\). 
Importantly, this symbolic distance is not injected into the loss function nor used to modify the internal attention layers. 
In principle, distance-based transitions can be verbalized such as \texttt{``strong upward shift is observed''}, 
However, we empirically found that providing only the compact symbolic pattern was sufficient in this work.
So our implementation injects the pattern string while using transition and periodicity cues only internally.

These distance-derived descriptors act as external inductive biases that guide how symbolic patterns are constructed, helping the SLM focus on large temporal shifts and periodic structures during prediction without modifying its internal architecture.

\subsubsection{Transition-Based Weighting}

Following the symbolization step, STM computes the symbolic variation:
\begin{equation}
\Delta_t = w(s_{t+1}) - w(s_t),
\label{t}
\end{equation}
where \(s_t\) denotes the encoded symbol at time step \(t\). 
Positive \(\Delta_t\) corresponds to an upward transition, whereas negative values indicate downward transitions.

Larger \(|\Delta_t|\) represent sharper changes in the time series, and STM leverages these transitions to highlight meaningful structural behaviors, consistent with the transition-based block in Fig.~\ref{stm}. 
The variation sequence \([\Delta_t]\) is subsequently used to identify change points and construct interpretable summaries, which are later injected into the prompt as contextual cues.

\subsubsection{Periodicity Detection}

To capture periodic or repeating patterns in temporal variations, STM examines the sequence \(\Delta_t\) and searches for a minimum repeating unit of length \(T\) that satisfies:
\begin{equation}
[\Delta_0, \Delta_1, \dots, \Delta_{T-1}] \approx 
[\Delta_T, \Delta_{T+1}, \dots, \Delta_{2T-1}].
\label{Periodicity Detection}
\end{equation}

Once detected, STM assigns periodic weights \(p(T)\) to emphasize transitions consistent with seasonal or cyclic behavior. 
This corresponds to the periodicity detection branch in Fig.~\ref{stm}, where symbolic transitions and repeated patterns are integrated.

Although simple in form, this periodicity detection is intentionally lightweight to support efficient execution in reprogramming pipelines for small language models. 
The detected pattern and its periodic signature are later embedded into the prompt as symbolic descriptors.

\subsection{Attention Score}

Unlike conventional QKV-based attention such as Eq.~(\ref{a}), STM computes an external attention score \(\alpha_t\) using symbolic transitions:
\begin{equation}
\alpha_t = \left(1 - \frac{d(s_t, s_{t-1})}{D} \right) \cdot |\Delta_t| \cdot p(T_t),
\label{attention_scoring}
\end{equation}
where \(D\) is the maximum symbol distance. 
This symbolic attention is not used to adjust internal model weights but instead serves as a descriptive weighting that highlights salient time steps.

Directional consistency is optionally enhanced when \(\text{sign}(\Delta_t)\) aligns with the overall trend. 
Furthermore, periodic positions receive additional emphasis. 
All attention scores are normalized to ensure \(\sum_t \alpha_t = 1\).

In line with the reprogramming pipeline of Fig.~\ref{stm}, STM computes symbolic transitions, periodic cues, and attention scores internally.
The final attention score $\alpha_t$ is used to construct a temporally weighted summary of the input sequence.
In our implementation, the numeric values of $\alpha_t$ are not directly exposed to the model; instead, their effects are implicitly reflected through the symbolic pattern injected into the prompt, while $\alpha_t$ itself  remains an internal descriptor.
Therefore, models receive concise and temporally informed representations that leverage structured symbolic information to guide next-step predictions without modifying model parameters.

\subsection{Model Integration}

STM is implemented as a lightweight wrapper module around input sequences, as shown in Fig.~\ref{overview}. STM does not modify model weights nor internal attention, ensuring full compatibility with device-constrained SLMs.
Rather than modifying the backbone model, STM symbolically re-encodes the input time series, computes transition- and periodicity-based descriptors, and integrates them into the prompt as auxiliary context.
This wrapper-based approach provides compatibility across diverse small language models while improving predictive performance through symbol-driven contextual reasoning.

\section{Experiments}
\label{experiments}

In this section, we demonstrate the effectiveness of our proposed STM using various SLMs as backbone models.
We detail the workflow in which STM is applied to LM using different kinds of time series data used in academia and industry for experiments.

In doing so, we demonstrate our work by evaluating the performance before and after the application of the proposed STM for different SLMs.
In addition, in this work, we analyze the correlation between the performance improvements of different SLMs and their architectures through STM-applied experiments and derive insights into future research directions.

\subsection{Data Acquisition and Pre-Processing}
\label{data_acquisition}

\subsubsection{Dataset Description}

\begin{figure}[t]
  \centering
  \begin{subfigure}[b]{0.48\linewidth}
    \centering
    \includegraphics[width=\linewidth]{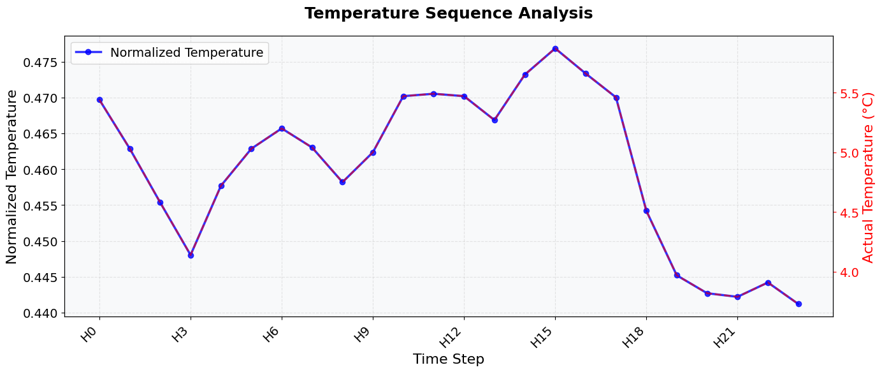}
    \caption{Temperature data of climate dataset (original).}
    \Description{Original temperature time-series from the climate dataset.}
    \label{temperature_original}
  \end{subfigure}
  \hfill
  \begin{subfigure}[b]{0.48\linewidth}
    \centering
    \includegraphics[width=\linewidth]{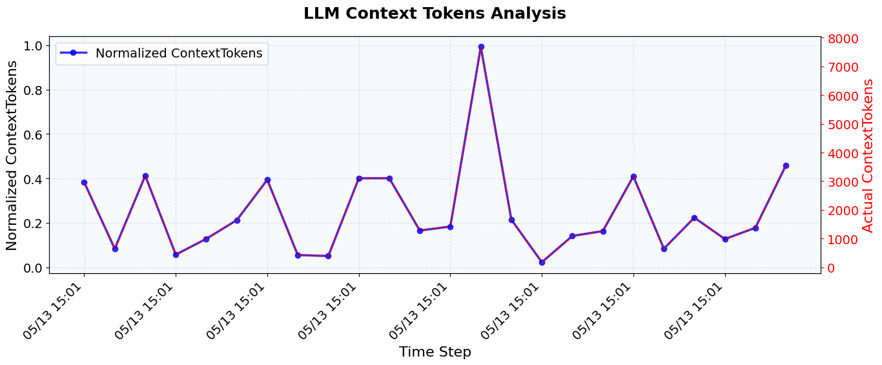}
    \caption{LLM traffic data (original code context tokens).}
    \Description{Original LLM traffic time-series represented as code context tokens.}
    \label{code_original}
  \end{subfigure}

  \vspace{0.7em}

  \begin{subfigure}[b]{0.48\linewidth}
    \centering
    \includegraphics[width=\linewidth]{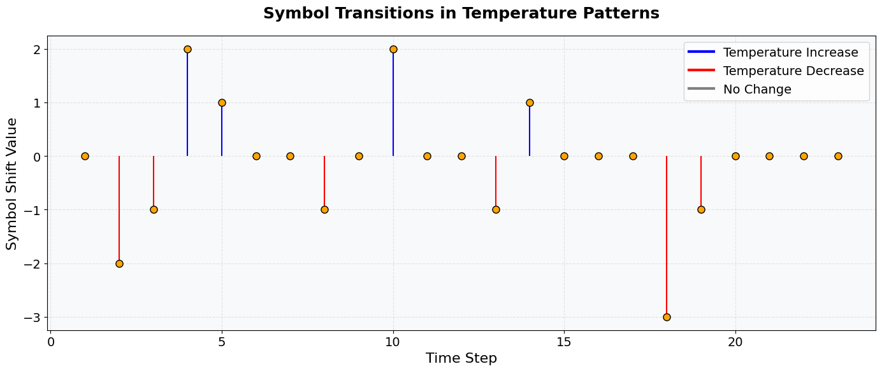}
    \caption{Temperature data after symbolic coding.}
    \Description{Symbolic-coded temperature time-series.}
    \label{temperature_symbol}
  \end{subfigure}
  \hfill
  \begin{subfigure}[b]{0.48\linewidth}
    \centering
    \includegraphics[width=\linewidth]{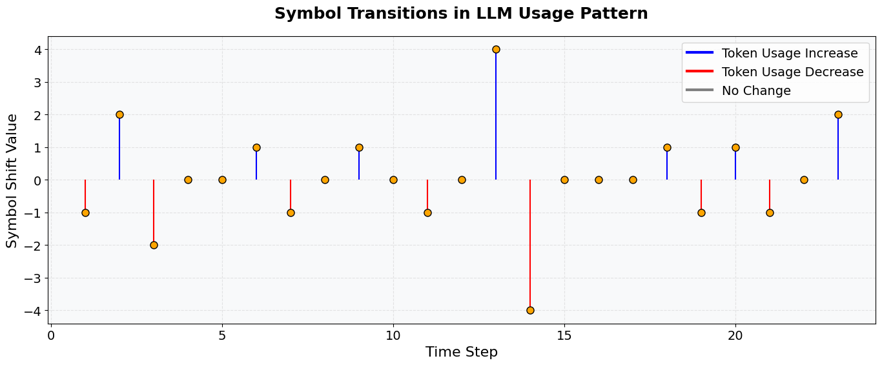}
    \caption{LLM traffic data after symbolic coding.}
    \Description{Symbolic-coded LLM traffic time-series.}
    \label{code_symbol}
  \end{subfigure}

  \vspace{0.7em}

  \begin{subfigure}[b]{0.48\linewidth}
    \centering
    \includegraphics[width=\linewidth]{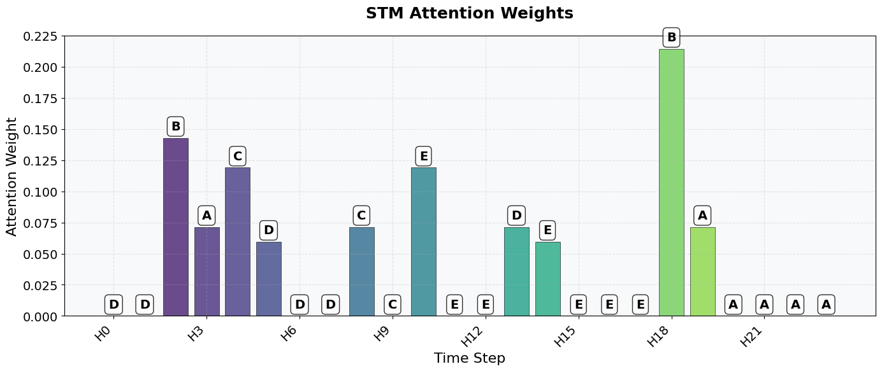}
    \caption{STM-weighted symbolic temperature data.}
    \Description{STM-weighted symbolic representation of temperature time-series.}
    \label{temperature_sttm}
  \end{subfigure}
  \hfill
  \begin{subfigure}[b]{0.48\linewidth}
    \centering
    \includegraphics[width=\linewidth]{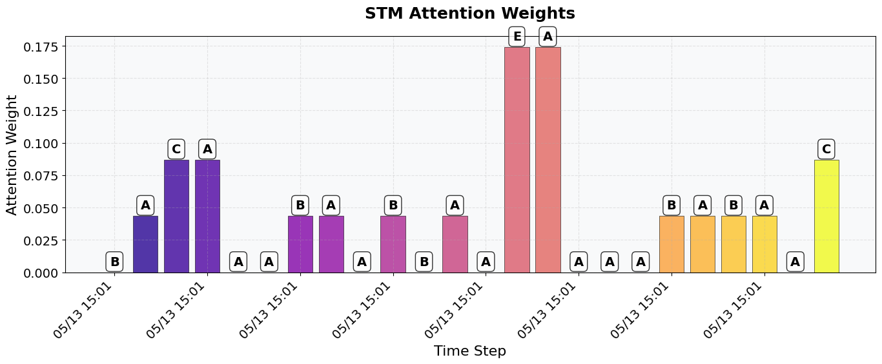}
    \caption{STM-weighted symbolic LLM traffic data.}
    \Description{STM-weighted symbolic representation of LLM traffic time-series.}
    \label{code_sttm}
  \end{subfigure}

  \caption{STM visualizations for climate temperature and LLM traffic datasets: 
  (a,b) original time series, (c,d) symbolic-coded representations, and (e,f) STM-weighted symbolic encodings.}
  \Description{STM visualization for climate temperature and LLM traffic datasets, showing original, symbolic-coded, and STM-weighted symbolic representations in a 3 by 2 layout.}
  \label{fig:stm_visualization}
\end{figure}

In this work, we prepare time series data from various disciplines as follows to closely verify the effectiveness of the proposed STM:

\begin{itemize}
    \item \textbf{Climate time series data:} We utilize the Max-Planck Jena Climate dataset, which provides temperature measurements collected at 10-minute intervals as shown in Fig. \ref{temperature_original} \cite{jena_climate_2009_2016}. For practical forecasting horizons, we resample these data to hourly intervals, resulting in sequences of hourly temperature.

    \item \textbf{LLM traffic time series data:} We obtained and experimented with a dataset that recorded a week of ContextTokens logs called at every timestamp for Azure LLM services in the Microsoft Azure data center as shown in Fig. \ref{code_original} \cite{azure_llm_inference_2023}.
    
\end{itemize}

Our work builds upon the \textit{Time-LLM} framework proposed in \cite{jin2023time}, which converts numerical time series signals into textual prompts for SLM. 
In detail, we used the Azure LLM dataset as it was, while using the Max-Planck Jena Climate dataset to ensure sufficient temporal granularity for the prediction task.

We use both datasets for training, validation, and test segmentation. The initial portion (70\%) is used for training, followed by a validation set (15\%) for hyperparameter tuning and a final test set (15\%) for out-of-sample performance evaluation. Each split preserves the temporal ordering to mirror real-world forecasting conditions.

\begin{table*}[t]
\centering
\caption{Summary of SLMs used for STM evaluation.}
\Description{Summary of SLMs used for STM evaluation.}
\label{slm_summary}

\resizebox{\textwidth}{!}{%
\begin{tabular}{l|c|c|c|c}
\hline
\textbf{Model} & \textbf{Params} & \textbf{Token} &
\textbf{Key characteristics} & \textbf{Distributor} \\
\hline
\texttt{Phi-3.5-mini}   & 3.8B & 128K &
Tunes through lightweight domain adaptation & Microsoft \\
\texttt{LLaMA-3.2-1B}   & 1.0B & 128K &
Rotary + grouped attention for on-device optimized & Meta \\
\texttt{DeepSeek-1.3B}  & 1.3B & 16K &
Pretraining on 2T tokens (87\% code, 13\% natural language) & DeepSeek \\
\texttt{Gemma-2B-it}    & 2.0B & 8K &
Rotary + multi-query attention for safe factual outputs & Google \\
\hline
\end{tabular}
} 

\end{table*}

\subsection{Implementation SLM with STM}
\label{implementation_slm}

Based on the previously acquired time series data, the STM framework begins with two steps.
First, we quantize the preprocessed time series data for the implementation of SLM with STM.
In these Fig.~\ref{temperature_symbol} and Fig.~\ref{code_symbol}, we derive symbol shift sequences by calculating differences between quantized symbols to show which changes STM values.

\subsubsection{Prompt Construction}
We convert SLM-interpreted temperature and network traffic data to text prompts below by referring to the existing studied prompt engineering \cite{jin2023time}:
To implement various SLMs with STM using these symbol shift sequences, we construct a basic prompt before STM compatible with quantized time series data as follows:

\begin{itemize}
    \item \textbf{Temperature Forecasting via Prompting:}
\begin{quote}
\texttt{The temperature readings for the past 24 hours are: \(sequence_str\), What is the next temperature reading?}
\end{quote}

    \item \textbf{LLM Traffic Forecasting via Prompting:}

\begin{quote}
\texttt{Given the following sequence of normalized inference traffic values:
\(sequence_str\), Predict the next traffic value based on pattern and trends.}
\end{quote}

\end{itemize}
In the STM implementation, which is added to the above base prompt, each prompt contains a normalized numerical sequence with a light symbolic pattern generated by the STM.
Additionally, STM calculates transition sizes, periodic cues, and attention scores internally according to Eqs.~(\ref{t})--(\ref{attention_scoring}), and these internal computations are used only to interpret the sequence internally, while the prompt exposes only the final symbolic pattern.

In all cases, STM-generated symbolic patterns are attached as optional auxiliary context, while the underlying backbone model remains unchanged. 
This unified prompting scheme allows SLMs to generalize across domains without revealing task-specific engineering details.
This maintains the efficiency of the reprogramming approach and prevents modification of the backbone model.

\subsubsection{Operation of STM}

Alongside the textual representation of numerical values, STM computes transition- and periodicity-based descriptors using Eqs.~(\ref{t})--(\ref{attention_scoring}). 
Given an input sequence, STM proceeds as follows:

\begin{enumerate}
    \item \textbf{Symbolization:} The continuous time series is discretized into symbols \(\{s_t\}\) using the five-level encoding described in Section~\ref{design_principle_approach}.

    \item \textbf{Transition and Periodicity Scoring:} For each time step, STM computes symbolic variations \(\Delta_t\), detects periodic structure, and derives attention scores \(\{\alpha_t\}\) according to Eq.~(\ref{attention_scoring}). These internal computations characterize trend, level shifts, and seasonal structure.

    \item \textbf{Internal Summary Representation:} The symbolic transitions, periodic cues, and attention sizes calculated above are only used internally for sequence interpretation and are inserted with symbol patterns derived from raw sequences as indirect descriptions at prompts.

\end{enumerate}

STM preserves the structure inside the SLM to ensure its versatility across different environments and models. For this purpose, symbolic patterns are generated independently from the internal transition and periodicity computations, and serve as lightweight textual summaries attached to the prompt.

Therefore, STM acts as a cognition-inspired external mechanism that lightly augments the input description without modifying the backbone model.
This design maintains compatibility among heterogeneous SLMs, and enables effective symbol-based contextual reasoning across heterogeneous SLMs.

\begin{table}
\centering
\caption{Parameters for STM-based SLMs}
\Description{Parameters for STM-based SLMs}
\begin{tabularx}{\linewidth}{X|X}
\hline
\textbf{Parameter} & \textbf{Value} \\
\hline
\texttt{num\_return\_sequences}  & $1$ \\
\texttt{temperature} & $0.12$ \\
\texttt{top\_p} & $0.9$ \\
\texttt{do\_sample} & \texttt{False} \\
\hline
\end{tabularx}
\label{Parameters}
\end{table}

\label{performance_evaluations}

\begin{table*}
\centering
\caption{Performance comparison of base models and STM-enhanced models across time-series forecasting tasks. Specific values are representative samples, while MAE and MSE are averaged over the test set.}
\Description{Performance comparison of base models and STM-enhanced models across time-series forecasting tasks. Specific values are representative samples, while MAE and MSE are averaged over the test set.}
\label{performance_evaluation_actual}
\renewcommand{\arraystretch}{1.2}
\resizebox{\linewidth}{!}{%
\begin{tabular}{c|c||r|r|r||r|r|r||r|r|r}
\toprule
\multirow{2}{*}{\textbf{Model}} &
\multirow{2}{*}{\textbf{Task}} &
\multicolumn{3}{c||}{\textbf{A representative example}} &
\multicolumn{3}{c||}{\textbf{MAE metric}} &
\multicolumn{3}{c}{\textbf{MSE metric}} \\
\cmidrule{3-11}
& & \textbf{Ground Truth} & \textbf{Base Pred} & \textbf{STM Pred} &
  \textbf{MAE (Base)} & \textbf{MAE (STM)} & \textbf{MAE improv. (\%)} &
  \textbf{MSE (Base)} & \textbf{MSE (STM)} & \textbf{MSE improv. (\%)} \\
\midrule
\texttt{Phi}      & Temperature & 12.85~°C & 11.64~°C & 12.48~°C & 1.2090 & 0.4459 & \textbf{63.12} & 3.6740 & 0.3648 & \textbf{90.07} \\
\texttt{Phi}      & LLM traffic & 273M      & 796M     & 347M     & 2.48e+08 & 1.12e+08 & \textbf{54.64} & 1.067e+17 & 2.072e+16 & \textbf{80.58} \\
\texttt{LLaMA}    & Temperature & 12.63~°C & 17.17~°C & 12.42~°C & 3.4022 & 1.9679 & \textbf{42.15} & 15.0385 & 8.3692 & \textbf{44.33} \\
\texttt{LLaMA}    & LLM traffic & 273M      & 796M     & 347M     & 3.70e+08 & 1.12e+08 & \textbf{69.73} & 1.797e+17 & 2.072e+16 & \textbf{88.47} \\
\texttt{DeepSeek} & Temperature & 12.48~°C & 12.24~°C & 11.64~°C & 1.9611 & 0.9536 & \textbf{51.37} & 5.0441 & 1.2290 & \textbf{75.63} \\
\texttt{DeepSeek} & LLM traffic & 273M      & 258M     & 181M     & 3.59e+08 & 3.22e+08 & 10.35        & 1.668e+17 & 1.335e+17 & 19.96        \\
\texttt{Gemma}    & Temperature & 12.81~°C & 16.81~°C & 17.35~°C & 4.5017 & 4.4537 & 1.07          & 21.4292 & 20.7319 & 3.25          \\
\texttt{Gemma}    & LLM traffic & 273M      & 66M      & 103M     & 3.57e+08 & 3.21e+08 & 10.06        & 1.829e+17 & 1.592e+17 & 12.94        \\
\bottomrule
\end{tabular}%
}
\end{table*}

\subsubsection{Model Configuration and Training}
We constructed an inference model with STM applied to evaluate the performance of STM based on different SLMs, as shown in Fig.~\ref{overview}.
To verify the performance improvement effect and generalizability of the proposed STM, we selected SLMs with various purposes and conditions as Tables~\ref{slm_summary} \cite{lu2024small}.

Each SLM model is configured for inference with the parameters as Tables~\ref{Parameters} for a consistent comparison of the collected time series data. STM operates purely through prompt reprogramming at the point of inference and no additional fine-tuning is performed

Specifically, we set the number of return sequences of SLMs with each STM to 1, ensuring deterministic output and reproducibility.
Additionally, we disabled sampling by setting \texttt{do\_sample = False}, thus enforcing greedy decoding, which allows for direct measurement of the impact of STM on model behavior without stochastic variability.  
To further suppress the randomness of token selection, a \texttt{temperature} value of 0.12 was used to promote a stable predictive path.
Furthermore, we set \texttt{top\_p} to 0.9 to maintain the coverage of the output.

These configurations were chosen to isolate the symbolic attention-weighting effect based on STM rather than relying on generational diversity or sampling heuristics, based on parameter settings previously studied \cite{yao2024survey, xu2018pressure, song2024good}.  
In this evaluation, parameter settings are equally adjusted for fair comparison across all time series data experiments.
These parameter values are applied to SLM-based inference models with several STMs such as Phi, DeepSeek, LLama, and Gemma.


While encoder-based models have shown strong performance in tasks such as classification and sentence-level inference, we select the above decoder-based models for time series prediction tasks where autoregressive structures are important \cite{podkorytov2021can}.

\subsection{Performance Evaluations}

We represent the effects of STM on various SLMs as above through quantitative comparisons of STM-enhanced models with the performance of the underlying model, as shown in Table~\ref{performance_evaluation_actual}.

In both time series data domains, STM mostly clearly improves the inference prediction error for three of the four models. In \texttt{Phi-3.5-mini}, STM reduces MAE by \textbf{63.1\%} for temperature and \textbf{54.6\%} for LLM traffic data, while MSE drops even more drastically by \textbf{90.1\%} and \textbf{80.6\%}, respectively.

\texttt{LLaMA-3.2-1B} shows the greatest absolute gain in traffic, with MAE reduced by \textbf{69.7\%} and MSE by \textbf{88.5\%}, while temperature also improves by \textbf{42.1\%} MAE and \textbf{44.3\%} MSE.
\texttt{DeepSeek-1.3B} shows a meaningful improvement of \textbf{51.4\%} in temperature MAE and \textbf{75.6\%} in MSE, as well as a moderate gain of \textbf{10.4\%} MAE and \textbf{20.0\%} MSE in LLM traffic data.
Finally, \texttt{Gemma-2B-it} combines multiple attention mechanisms and rotational position embeddings with relatively small context windows. It encodes part of the structure supplied by the STM, indicating that the incremental benefit is limited, with only \textbf{1.1\%} to \textbf{13.0\%} gain depending on the domain and metric.

To complement the analysis presented in Sections~\ref{symbol}, additional ablation was performed using seven symbolic levels.
The 7-symbol configuration also showed performance improvement over non-symbolic baselines, but the overall accuracy was, on average, approximately 10\% lower than the 5-symbol design. This further proves that over-segmentation reduces the symbolic efficiency of STM, confirming that 5-level representation is the most balanced and robust choice.

Furthermore, we measured GPU memory and latency in the RTX 5060ti environment. Through runtime profiling, we find that applying STM increases the maximum GPU-enabled peak memory by 0.06\% with a delay overhead of only 0.64\%. Although not an actual embedded board, the overhead is negligibly small.
This result shows a significant improvement in prediction accuracy of low-burden SLMs in environments with limited time inference, such as IoT environments.

\begin{figure}
\centering \makeatletter\IfFileExists{fig/corr.png}{\includegraphics[width=0.9\textwidth]{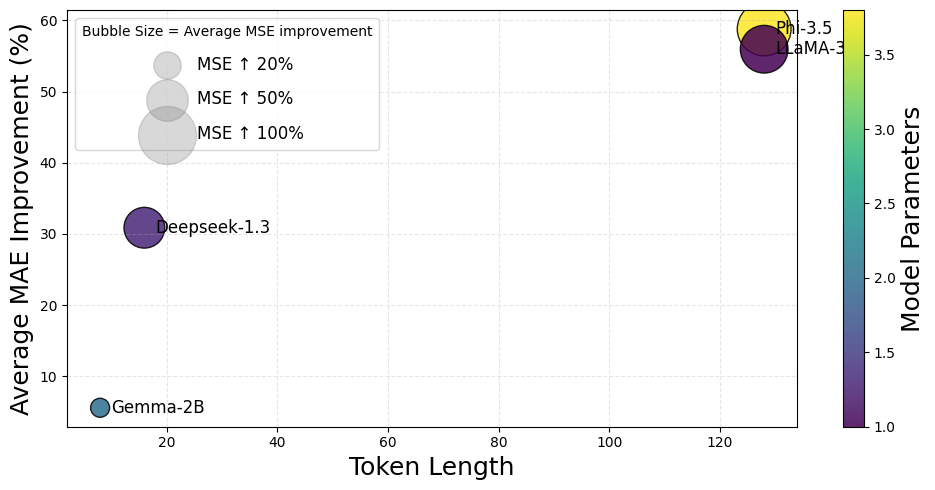}}{}
\makeatother 
\caption{{Effectiveness of STM on each model.}}
\Description{{Effectiveness of STM on each model.}}
\label{corr}
\end{figure}

We summarize the correlation of the effect of STM on the context and token features of each model organized in Table~\ref{slm_summary} as shown in Fig.~\ref{corr}.

Models such as \texttt{Phi-3.5-mini} and \texttt{LLaMA-3.2-1B} with long context windows, as shown in  Fig.~\ref{corr}, keep more timestep information in one prompt, suggesting the greatest performance gain by the weights emphasized by STM.
Furthermore, we find that larger parameters proportional to inference space and expressive power digest the symbolic information provided by STM better. This not only reduces the absolute error exhibited by MAE, but also shows that STM well mitigates sharp level fluctuation outliers through MSE.

However, the results of \texttt{Gemma-2B-it} show that the effect of STM is minimal if the context is short, even if the parameters are large.
\texttt{Phi-3.5-mini} and \texttt{LLaMA-3.2-1B} are \texttt{text-only model} such that \textbf{128 K-token context} provided by each model is used by following common commands. Thus, we suggest that STM gets the greatest error reduction when injecting trend-move-periodicity importance at the prompt layer.

In contrast, \texttt{Gemma-2B-it}, despite its relatively large parameter size, exhibits only marginal improvement with STM. This is likely due to its short context window (8K), which limits its ability to benefit from symbolic temporal transitions.  It encodes part of the structure supplied by the STM, indicating that the incremental benefit is up to 10\%.
Additionally, \texttt{DeepSeek-coder-1.3B} already tokenizes 87\% code text as a syntax-level symbol, indicating that its underlying representation is symbolic, thus indicating that STM is effective in smooth time series, as shown in Fig.~\ref{temperature_original}.
We also show that it still helps with traffic data with periodic bursts, as shown in Fig.~\ref{code_original}.

However, we find that \texttt{DeepSeek-coder-1.3B}, which has a lot of pre-learning, lacks parameters and tokens compared to models such as \texttt{Phi-3.5-mini} and \texttt{LLaMA-3.2-1B}, limiting performance improvements.
Finally, \texttt{Gemma-2B-it} uses rotation position embeddings and multiple attention mechanisms and rotation position embeddings, with a relatively small 8K token context window. This encodes part of the structure provided by STM, indicating that the incremental advantage is up to 10\%.

In conclusion, STM demonstrates maximal effectiveness when both sufficient context length and parameter capacity are present. Its marginal utility decreases in models with inherently symbolic or domain-specific representations, such as \texttt{DeepSeek-1.3B} and \texttt{Gemma-2B-it}, where structural priors are already embedded through architecture or pretraining data.
Therefore, we propose that STM has marginal utility for models with domain-specific pre-learning encoding and embedding structures.

\subsection{Discussion}
Our results show the potential of STM to enhance different SLMs. However, further discussion and research are needed to determine whether the five-level symbols presented in this paper are appropriate categories. In doing so, we plan to investigate how symbol diversity interacts with features such as sequence noise, periodicity strength, and model depth.

Furthermore, the symbolic representation is expected to improve interpretability, allowing practitioners to track the focus of model on critical temperature fluctuations and iterative cycles. In future studies, we plan to further improve the robustness by incorporating performance enhancements into more diverse irregular or multivariate time series areas such as finance or healthcare.

\section{Conclusion}
In this paper, we introduce the Symbolic Transition Mechanism (STM), a lightweight module that provides explicit temporal significance cues to improve the predictive power of language models.
Focusing on symbol quantization, transition weight estimation, and periodicity detection, STM is designed to work efficiently without modifying backbone parameters in small language models with resource constraints.

Across temperature and network traffic datasets, STM reduced MAE by up to 69\% and MSE by up to 90\% compared to baseline SLMs. Furthermore, the additional computational burden measured in GPU peak memory and inference latency is minimized, demonstrating practical deployability in edge-oriented devices.

Moreover, our ablation analysis shows that there is a stable cognitive correlation between the number of symbolic levels and model performance. In particular, we find that a five-level compact symbolic space provides an optimal balance between segmentation and generalization of various LMs.

Future work will extend STM to richer symbolic representations, broader time series domains, and integration with larger LMs beyond the SLMs considered in this study. We believe that STM provides a practical pathway to improve time series prediction in both resource-limited deployments and universal AI systems.

\begin{acks}
This work was supported by the Korea Institute of Energy Technology Evaluation and Planning (KETEP) and the Ministry of Climate, Energy \& Environment(MCEE) of the Republic of Korea (RS-2022-KP002860).
This research was supported by the MSIT (Ministry of Science and ICT), Korea, under the ITRC (Information Technology Research Center) support program (IITP-2021-0-01835) supervised by the IITP (Institute of Information \& Communications Technology Planning \& Evaluation).
\end{acks}

\bibliography{samples/article}

\end{document}